\documentclass{article}

\usepackage{times}
\usepackage{graphicx} 
\usepackage[caption=false]{subfig}
\usepackage{amssymb}

\usepackage{natbib}

\usepackage{algorithm}
\usepackage{algorithmic}
\usepackage{enumitem}
\usepackage{multirow}
\usepackage{multicol}
\usepackage{amsthm}
\usepackage{stmaryrd}
\usepackage{xcolor}
\usepackage{arydshln}

\setlist{nosep}

\usepackage{float}

\usepackage{amsfonts}

\definecolor{darkgreen}{rgb}{0,0.4,0}

\def\xx{%
\raggedright
\def\UTFviii@two@octets##1\x##2{\expandafter
    \UTFviii@defined\csname u8:##1\string##2\endcsname\x}%
\def\UTFviii@three@octets##1\x##2##3{\expandafter
    \UTFviii@defined\csname u8:##1\string##2\string##3\endcsname\x}
\def\stopx\x{}%
}

\newcommand{\hyphen}{- }
\newcommand{\uscore}{\char`_}

\newcommand*{\shifttext}[2]{%
  \settowidth{\@tempdima}{#2}%
  \makebox[\@tempdima]{\hspace*{#1}#2}%
}
\newcommand{\shiftleft}[2]{\makebox[0pt][r]{\makebox[#1][l]{#2}}}
\makeatother

\usepackage{hyperref}

\usepackage[accepted]{icml2017}

\usepackage{amsfonts}

\icmltitlerunning{RobustFill: Neural Program Learning under Noisy I/O}
\begin{document} 

\twocolumn[
\icmltitle{RobustFill: Neural Program Learning under Noisy I/O}


\icmlsetsymbol{equal}{*}

\begin{icmlauthorlist}
\icmlauthor{Jacob Devlin}{equal,msr}
\icmlauthor{Jonathan Uesato}{equal,mit}
\icmlauthor{Surya Bhupatiraju}{equal,mit}
\icmlauthor{Rishabh Singh}{msr}
\icmlauthor{Abdel-rahman Mohamed}{msr}
\icmlauthor{Pushmeet Kohli}{msr}
\end{icmlauthorlist}

\icmlaffiliation{msr}{Microsoft Research, Redmond, Washington, USA}
\icmlaffiliation{mit}{MIT, New London, Cambridge, Massachusetts, USA}

\icmlcorrespondingauthor{Jacob Devlin}{jdevlin@microsoft.com}

\icmlkeywords{deep learning, program synthesis, program induction, program learning, attention, noisy data, machine learning, ICML}

\vskip 0.3in
]

\printAffiliationsAndNotice{\icmlEqualContribution} 

\begin{abstract} 
The problem of automatically generating a computer program from some specification has been studied since the early days of AI. 
Recently, two competing approaches for {\it automatic program learning} have received significant attention: (1) {\it neural program synthesis}, where a neural network is conditioned on input/output (I/O) examples and learns to generate a program, and (2) {\it neural program induction}, where a neural network generates new outputs directly using a {\it latent} program representation.

Here, for the first time, we directly compare both approaches on a large-scale, real-world learning task. We additionally contrast to 
{\it rule-based program synthesis}, which uses hand-crafted semantics to guide the program generation. 
Our neural models use a modified attention RNN to allow encoding of variable-sized sets of I/O pairs.
Our best synthesis model achieves 92\% accuracy on a real-world test set, compared to the 34\% accuracy of the previous best neural synthesis approach. The synthesis model also outperforms a comparable induction model on this task, but we more importantly demonstrate that the strength of each approach is highly dependent on the evaluation metric and end-user application. Finally, we show that we can train our neural models to remain very robust to the type of noise expected in real-world data (e.g., typos), while a highly-engineered rule-based system fails entirely.
\end{abstract} 

\section{Introduction}

The problem of \emph{program learning}, i.e. generating a program consistent with some specification, is one of the oldest problems in machine learning and artificial intelligence \citet{waldinger1969,manna1975}. The classical approach has been that of {\it rule-based program synthesis} \cite{manna1980}, where a formal grammar is used to derive a program from a well-defined specification. Providing a formal specification is often more difficult than writing the program itself, so modern program synthesis methods generally rely on {\it input/output examples} (I/O examples) to act as an approximate specification. Modern rule-based synthesis methods are typically centered around hand-crafted function semantics and pruning rules to search for programs consistent with the I/O examples \cite{gulwani2012,alur2013}.

\begin{figure}[t]
\begin{center}
\resizebox{0.37\textwidth}{!}{ 
\begin{tabular}{|l|l|}
\hline
\multicolumn{1}{|c|}{\textbf{Input String}} & \multicolumn{1}{|c|}{\textbf{Output String}} \\ \hline
john Smith & Smith, Jhn \\
DOUG Q. Macklin\quad \quad & Macklin, Doug \\
Frank Lee (123) & LEe, Frank \\
Laura Jane Jones & Jones, Laura \\ \hline
Steve P. Green (9) & ? \\ \hline
\multicolumn{2}{|c|}{\textbf{Program}} \\ \hline
\multicolumn{2}{|l|}{\small\tt{GetToken(Alpha, -1) $|$\ `,'\ $|$ \ ` '\ $|$}} \\
\multicolumn{2}{|l|}{\small\tt{ToCase(Proper, GetToken(Alpha, 1))}} \\
\hline
\end{tabular}
}
\end{center}
\caption{An anonymized example from FlashFillTest with noise (typos). The goal of the task is to fill in the blank (i.e., `?' = `Green, Steve'). Synthesis approaches achieve this by generating a program like the one shown. Induction approaches generate the new output string directly, conditioned on the the other examples.}
\label{fig:main_example}
\vspace{-12pt}
\end{figure}
These hand-engineered systems are often difficult to extend and fragile to noise, so {\it statistical program learning} methods have recently gained popularity, with a particular focus on neural network models. This work has fallen into two overarching categories: (1) {\it neural program synthesis}, where the program is generated by a neural network conditioned on the I/O examples \cite{deepcoder,parisotto2017,terpret,forth}, and (2) {\it neural program induction}, where network learns to generate the output directly using a {\it latent} program representation \cite{ntm, dnc,nram,ngpu,stackrnn,npi, neuralprogrammer}. Although many of these papers have achieved impressive results on a variety of tasks, none have thoroughly compared induction and synthesis approaches on a real-world test set. In this work, we not only demonstrate strong empirical results compared to past work, we also directly contrast the strengths and weaknesses of both {\it neural program learning} approaches for the first time.



The primary task evaluated for this work is a Programming By Example (PBE) system for string transformations similar to {\it FlashFill} \cite{gulwani2012,gulwani2011}. FlashFill allows Microsoft Excel end-users to perform regular expression-based string transformations using examples without having to write complex macros. For example, a user may want to extract zip codes from a text field containing  addresses, or transform a timestamp to a different format. An example is shown in Figure~\ref{fig:main_example}. A user manually provides a {\it small number} of example output strings to convey the desired intent and the goal of FlashFill is to generalize the examples to automatically generate the corresponding outputs for the remaining input strings. Since the end goal is to emit the correct output strings, and not a program, the task itself is agnostic to whether a  {\it synthesis} or {\it induction} approach is taken.

For modeling, we develop novel variants of the attentional RNN architecture~\cite{bahdanau2014} to encode a variable-length unordered set of input-output examples. For program representation, we have developed a domain-specific language (DSL), similar to that of \citet{gulwani2012}, that defines an expressive class of regular expression-based string transformations. The neural network is then used to generate a program in the DSL (for synthesis) or an output string (for induction). Both systems are trained end-to-end using a large set of input-output examples and programs uniformly sampled from the DSL.

We compare our neural induction model, neural synthesis model, and the rule-based architecture of \citet{gulwani2012} on a real-world FlashFill test set. We also inject varying amounts of noise (i.e., simulated typos) into the FlashFill test examples to model the robustness of different learning approaches. While the manual approaches work reasonably well for well-formed I/O examples, we show that its performance degrades dramatically in presence of even small amounts of noise. We show that our neural architectures are significantly more robust in presence of noise and moreover obtain an accuracy comparable to manual approaches even for non-noisy examples.

This paper makes the following key contributions:
\begin{itemize}
\item We present a novel variant of the attentional RNN architecture, which allows for encoding of a variable-size set of input-output examples. 
\item We evaluate the architecture on 205 real-world FlashFill instances and significantly outperform the previous best statistical system (92\% vs. 34\% accuracy).
\item We compare the model to a hand-crafted synthesis algorithm and show that while both systems achieve similar performance on clean test data, our model is significantly more robust to realistic noise (with noise, 80\% accuracy vs. 6\% accuracy).
\item We compare our neural synthesis architecture with a  neural induction architecture, and demonstrate that each approach has its own strengths under different evaluation metrics and decoding constraints.
\end{itemize}

\section{Related Work}
\label{sec:related_work}

There has been an abundance of recent work on neural program induction and synthesis. 

{\bf Neural Program Induction: } Neural Turing Machine (NTM)~\cite{ntm} uses a neural controller to read and write to an external memory tape using soft attention and is able to learn simple algorithmic tasks such as array copying and sorting. Stack-RNNs~\cite{stackrnn} augment a neural controller with an external stack-structured memory and is able to learn algorithmic patterns of small description length. Neural GPU~\cite{ngpu} presents a Turing-complete model similar to NTM, but with a parallel and shallow design similar to that of GPUs, and is able to learn complex algorithms such as long binary multiplication. Neural Programmer-Interpreters~\cite{npi} teach a controller to learn algorithms from program traces as opposed to examples. Neural Random-Access Machines~\cite{nram} uses a continuous representation of 14 high-level modules consisting of simple arithmetic functions and reading/writing to a variable-size random-access memory to learn algorithmic tasks requiring pointer manipulation and dereferencing to memory. The domain of string transformations is different than the domains handled by these approaches and moreover, unlike RobustFill, these approaches need to be re-trained per problem instance.


{\bf Neural Program Synthesis: } The most closely related work to ours uses a Recursive-Reverse-Recursive neural network (R3NN) to learn string transformation programs from examples \cite{parisotto2017}, and is directly compared in Section~\ref{sec:past_work}. DeepCoder~\cite{deepcoder} trains a neural network to predict a distribution over possible functions useful for a given task from input-output examples, which is used to augment an external search algorithm. Unlike DeepCoder, RobustFill performs an end-to-end synthesis of programs from examples. Terpret~\cite{terpret} and Neural Forth~\cite{forth} allow programmers to write sketches of partial programs to express prior procedural knowledge, which are then completed by training neural networks on examples.



{\bf DSL-based synthesis: } Non-statistical DSL-based synthesis approaches~\cite{gulwani2012} exploit independence properties of DSL operators to develop a divide-and-conquer based search algorithm with several hand-crafted pruning and ranking heuristics~\cite{flashmeta}. In this work, we present a neural architecture to automatically learn an efficient synthesis algorithm. There is also some work on using learnt clues to guide the search in DSL expansions~\cite{menon}, but this requires hand-coded textual features of examples.


\section{Problem Overview}

We now formally define the problem setting and the domain-specific language of string transformations.

\subsection{Problem Formulation}
\label{sec:con_vs_gen_desc}

Given a set of input-output (I/O) string examples $(I_1,O_1) , ... , (I_n,O_n)$, and a set of unpaired input strings $I^y_1, ..., I^y_m$, the goal of of this task is to generate the corresponding output strings, $O^y_1, ..., O^y_m$. For each example set, we assume there exists at least one program $P$ that will correctly transform all of these examples, i.e., $P(I_1) \rightarrow O_1, ...,  P(I^y_1) \rightarrow O^y_1, ...$ Throughout this work, we refer to $(I_j, O_j)$ as {\it observed examples} and $(I^y_j, O^y_j)$ as {\it assessment examples}. We use {\tt InStr} and {\tt OutStr} to generically refer to I/O examples that may be observed or assessment. We refer to this complete set of information as an {\it instance}:
\begin{center}
\begin{tabular}{|l|l|}
\hline
$I_1 =$ {\tt January} & $O_1 = $ {\tt jan} \\
$I_2 =$ {\tt February} & $O_2 = $ {\tt feb} \\ 
$I_3 =$ {\tt March} & $O_3 = $ {\tt mar} \\ \hline
$I^y_1 =$ {\tt April} & $O^y_1 = $ {\tt apr} \\
$I^y_2 =$ {\tt May} & $O^y_2 = $ {\tt may} \\ \hline
\multicolumn{2}{|c|}{$P=$ {\tt ToCase(Lower, SubStr(1,3))}} \\ 
\hline
\end{tabular}
\end{center}

Intuitively, imagine that a (non-programmer) user has a large list of {\tt InStr} which they wish to process in some way. The goal is to only require the user to manually create a small number of corresponding {\tt OutStr}, and the system will generate the remaining {\tt OutStr} automatically. 

In the {\it program synthesis} approach, we train a neural model which takes $(I_1,O_1) , ... , (I_n,O_n)$ as input and generates $P$ as output, token-by-token. It is trained fully supervised on a large corpus of synthetic I/O Example + Program pairs. It is {\it not} conditioned on the assessment input strings $I^y_j$, but it could be in future work. At test time, the model is provided with new set of observed I/O examples and attempts to generate the corresponding $P$ which it (maybe) has never seen in training. Crucially, the system can actually execute the generated $P$ on each {\it observed} input string $I_j$ and check if it produces $O_j$.\footnote{This execution is deterministic, not neural.}  If not, it knows that $P$ cannot be the correct program, and it can search for a different $P$. Of course, even if $P$ is {\it consistent} on all observed examples, there is no guarantee that it will {\it generalize} to new examples (i.e., assessment examples). We can think of {\it consistency} as a necessary, but not sufficient, condition. The actual success metric is whether this program {\it generalizes} to the corresponding assessment examples, i.e., $P(I^y_j) = O^y_j$. There also may be multiple valid programs.

In the {\it program induction} approach, we train a neural model which takes $(I_1,O_1) , ... , (I_n,O_n)$ and $I^y$ as input and generates $O^y$ as output, character-by-character. Our current model decodes each assessment example independently. Crucially, the induction model makes no explicit use use of program $P$ at training or test time. Instead, we say that it induces a {\it latent} representation of the program. If we had a large corpus of real-world I/O examples, we could in fact train an induction model without any explicit program representation. Since such a corpus is not available, it is trained on the same synthesized I/O Examples as the synthesis model. Note that since the program representation is latent, there is no way to measure consistency. 

We can comparably evaluate both approaches by measuring {\it generalization accuracy}, which is the percent of test instances for which the system has successfully produced the correct {\tt OutStr} for all assessment examples. For synthesis this means $P(I^y_j)=O^y_j$ $\;\forall (I^y_j,O^y_j)$. For induction this means all $O^y$ generated by the system are exactly correct. We typically use four observed examples and six assessment examples per test instance. All six must be exactly correct for the model to get credit.

\subsection{The Domain Specific Language}
\label{sec:dsl}

The Domain Specific Language (DSL) used here to represent $P$ models a rich set of string transformations based on substring extractions, string conversions, and constant strings. The DSL is similar to the DSL described in \citet{parisotto2017}, but is extended to include nested expressions, arbitrary constant strings, and a powerful regex-based substring extraction function. The syntax of the DSL is shown in Figure~\ref{dslsyntax} and the formal semantics are presented in the supplementary material.

\begin{figure}[h]
\vspace{-10pt}
\begin{center}
\resizebox{0.39\textwidth}{!}{ 
\begin{minipage}{\linewidth}
  \begin{eqnarray*}
  \mbox{Program } p & := & \tt{Concat}(e_1, e_2, e_3, ...) \nonumber \\
  \mbox{Expression } e & := & f \; | \; n \; | \; n_1(n_2) \; | \; n(f) \; | \; \tt{ConstStr}(c) \nonumber \\
  \mbox{Substring } f & := & \tt{SubStr}(k_1,k_2)\\
  & | & \tt{GetSpan}(r_1,i_1,y_1,r_2,i_2,y_2) \nonumber \\
  \mbox{Nesting } \tt{n} & := & \tt{GetToken}(t, i) \; | \; \tt{ToCase}(s) \\
  & | & \tt{Replace}(\delta_1,\delta_2) \; | \; \tt{Trim}() \nonumber \\
  & | & \tt{GetUpto}(r) \; | \; \tt{GetFrom}(r) \\
  & | & \tt{GetFirst}(t,i) \; | \; \tt{GetAll}(t) \\
  \mbox{Regex } r & := & t_1 \; | \; \cdots \; | \; t_n \; | \; \delta_1 \; | \; \cdots \; | \; \delta_m \nonumber \\
  \mbox{Type } t & := & \tt{Number} \; | \; \tt{Word} \; | \; \tt{Alphanum} \nonumber \\
  & | & \tt{AllCaps} \; | \; \tt{PropCase} \; | \; \tt{Lower} \nonumber \\
  & | & \tt{Digit} \; | \; \tt{Char} \nonumber \\
  \mbox{Case } s & := & \tt{Proper} \; | \; \tt{AllCaps} \; | \; \tt{Lower} \nonumber \\
  \mbox{Position } k & := & -100, -99, ..., 1, 2, ... , 100 \nonumber \\
  \mbox{Index } i & := & -5, -4, -3, -2, 1, 2, 3, 4, 5 \nonumber \\
  \mbox{Character } c & := & \tt{A-Z},\tt{a-z},\tt{0-9},\tt{!?,@...} \nonumber \\
  \mbox{Delimiter } \delta & := & \tt{\&,.?!@()[]\%\{\}/:;\$\#"'}\ \nonumber \\
  \mbox{Boundary } y & := & \tt{Start} \; | \; \tt{End} \nonumber
  \end{eqnarray*}

\end{minipage}
}
\end{center}
\vspace{-6pt}
\caption{Syntax of the string transformation DSL.}
\label{dslsyntax}
\vspace{-4pt}
\end{figure}

A program $P: \tt{string} \Rightarrow \tt{string}$ in the DSL takes as input a string and returns another string as output. The top-level operator in the DSL is the {\tt Concat} operator that concatenates a finite list of string expressions $e_i$. A string expression $e$ can either be a substring expression $f$, a nesting expression $n$, or a constant string expression. A substring expression can either be defined using two constant positions indices $k_1$ and $k_2$ (where negative indices denote positions from the right), or using the $\tt{GetSpan(r_1,i_i,y_1,r_2,i_2,y_2)}$ construct that returns the substring between the $i_1^{\tt{th}}$ occurrence of regex $r_1$ and the $i_2^{\tt{th}}$ occurrence of regex $r_2$, where $y_1$ and $y_2$ denotes either the start or end of the corresponding regex matches. The nesting expressions allow for further nested string transformations on top of the substring expressions allowing to extract $k^{\tt{th}}$ occurrence of certain regex, perform casing transformations, and replacing a delimiter with another delimiter. The notation  $e_1\ \vert\ e_2\ \vert\ ...$ is sometimes used as a shorthand for ${\tt Concat}(e_1, e_2, ...)$. The nesting and substring expressions take a string as input (implicitly as a lambda parameter). We sometimes refer expressions such as {\tt ToCase(Lower)(v)} as {\tt ToCase(Lower,v)}.

There are approximately 30 million unique string expressions $e$, which can be concatenated to create arbitrarily long programs. Any search method that does  not encode inverse function semantics (either by hand or with a statistical model) cannot prune partial expressions. Thus, even efficient techniques like  dynamic programming (DP) with black-box expression evaluation would still have to search over many millions of candidates.

\subsection{Training Data and Test Sets}
\label{sec:data_synthesis}
Since there are only a few hundred real-world FlashFill instances, the data used to train the neural networks was synthesized automatically. To do this, we use a strategy of random sampling and generation. First, we randomly sample programs from our DSL, up to a maximum length (10 expressions). Given a sampled program, we compute a simple set of heuristic requirements on the {\tt InStr} such that the program can be executed without throwing an exception. For example, if an expression in the program retrieves the 4th number, the {\tt InStr} must have at least 4 numbers.
Then, each {\tt InStr} is generated as a random sequence of ASCII characters, constrained to satisfy the requirements. The corresponding {\tt OutStr} is generated by executing the program on the {\tt InStr}. 

For evaluating the trained models, we use {\it FlashFillTest}, a set of 205 real-world examples collected from Microsoft Excel spreadsheets, and provided to us by the authors of \citet{gulwani2012} and \citet{parisotto2017}. Each FlashFillTest instance has ten I/O examples, of which the first four are used as observed examples and the remaining six are used as assessment examples.\footnote{In cases where less than 4 observed examples are used, only the 6 assessment examples are used to measure generalization.} 
Some examples of FlashFillTest instances are provided in the supplementary material. 
Intuitively, it is possible to generalize to a real-word test set using randomly synthesized training because the model is learning {\it function semantics}, rather than a particular data distribution.

\section{Program Synthesis Model Architecture}
\label{sec:model_arch}

We model program synthesis as a sequence-to-sequence generation task, along the lines of past work in machine translation \cite{bahdanau2014}, image captioning \cite{xu2015}, and program induction \cite{zaremba2014}. In the most general description, we encode the observed I/O using a series of recurrent neural networks (RNN), and generate $P$ using another RNN one token at a time. The key challenge here is that in typical sequence-to-sequence modeling, the input to the model is a single sequence. In this case, the input is a variable-length, unordered set of sequence pairs, where each pair (i.e., an I/O example) has an internal conditional dependency. We describe and evaluate several multi-attentional variants of the attentional RNN architecture \cite{bahdanau2014} to model this scenario. 

\subsection{Single-Example Representation}

We first consider a model which only takes a single observed example $(I, O)$ as input, and produces a program $P$ as output. Note that this model is {\it not} conditioned on the assessment input $I^y$. In all models described here, $P$ is generated using a sequential RNN, rather than a hierarchical RNN \cite{parisotto2017,tai2015}.\footnote{Even though the DSL does allow limited hierarchy, preliminary experiments indicated that using a hierarchical representation did not add enough value to justify the computational cost.} As demonstrated in \citet{vinyals2015}, sequential RNNs can be surprisingly strong at representing hierarchical structures.

\begin{figure}[ht]
\begin{centering}
\includegraphics[width=200px]{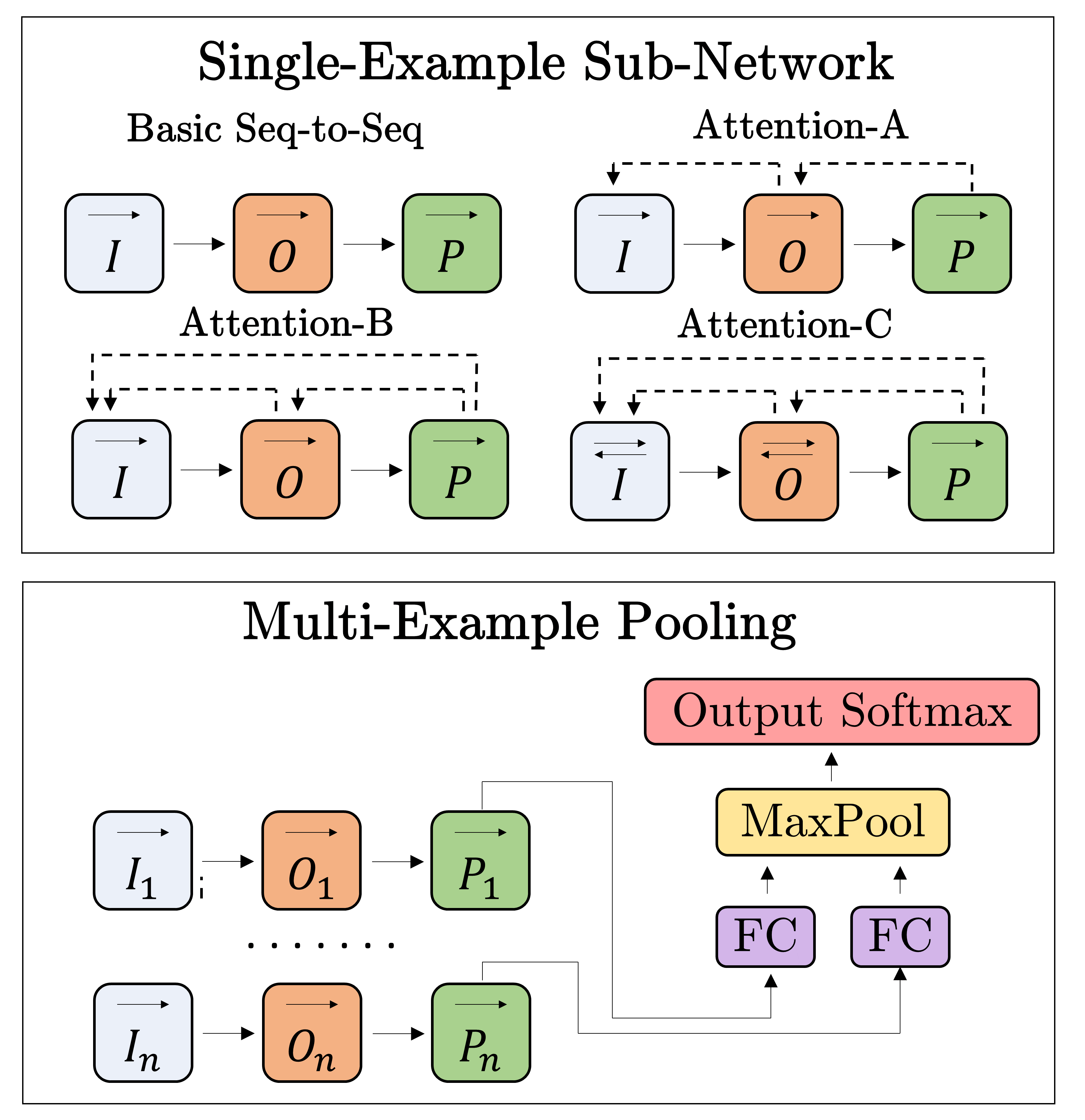}
\caption{The network architectures used for program synthesis. A dotted line from $x$ to $y$ means that $x$ attends to $y$.}
\label{fig:synthesis_arch_diagram}
\end{centering}
\vspace{-10pt}
\end{figure}

We explore four increasingly complex model architectures, shown visually in Figure~\ref{fig:synthesis_arch_diagram}:

\begin{itemize}
\item {\bf Basic Seq-to-Seq}: Each sequence is encoded with a non-attentional LSTM, and the final hidden state is used as the initial hidden state of the next LSTM.
\item {\bf Attention-A}: $O$ and $P$ are attentional LSTMs, with $O$ attending to $I$ and $P$ attending to $O$.\footnote{A variant where $O$ and $I$ are reversed performs significantly worse.}
\item {\bf Attention-B}: Same as Attention-A, but $P$ uses a {\it double attention} architecture, attending to both $O$ and $I$ simultaneously.
\item {\bf Attention-C}: Same as Attention-B, but $I$ and $O$ are bidirectional LSTMs.
\end{itemize}

In all cases, the {\tt InStr} and {\tt OutStr} are processed at the character level, so the input to $I$ and $O$ are character embeddings. The vocabulary consists of all 95 printable ASCII tokens.

The inputs and targets for the $P$ layer is the source-code-order linearization of the program. The vocabulary consists of 430 total program tokens, which includes all function names and parameter values, as well as special tokens for concatenation and end-of-sequence. Note that numerical parameters are also represented with embedding tokens. The model is trained to maximize the log-likelihood of the reference program $P$.

\subsection{Double Attention}
{
\setlength{\belowdisplayskip}{4pt} \setlength{\belowdisplayshortskip}{4pt}
\setlength{\abovedisplayskip}{4pt} \setlength{\abovedisplayshortskip}{4pt}
\textit{Double attention} is a straightforward extension to the standard attentional architecture, similar to the multimodal attention described in \citet{huang2016}. A typical attentional layer takes the following form:
\begin{eqnarray*}
s_i & = & Attention(h_{i-1}, x_i, S) \\
h_i & = & LSTM(h_{i-1}, x_i, s_i)
\end{eqnarray*}
Where $S$ is the set of vectors being attended to, $h_{i-1}$ is the previous recurrent state, and $x_i$ is the current input. The $Attention()$ function takes the form of the ``general'' model from \citet{luong2015}. Double attention takes the form:
\begin{eqnarray*}
s^A_i & = & Attention(h_{i-1}, x_i, S^A) \\
s^B_i & = & Attention(h_{i-1}, x_i, s^A_i, S^B) \\
h_i & = & LSTM(h_{i-1}, x_i, s^A_i, s^B_i)
\end{eqnarray*}
Note that $s^A_i$ is concatenated to $h_{i-1}$ when computing attention on $S^B$, so there is a directed dependence between the two attentions. Here, $S^A$ is $O$ and $S^B$ is $I$. In the LSTM, $s^A_i$ and $s^B_i$ are concatenated.
}

\subsection{Multi-Example Pooling}
\label{sec:att_seq}
The previous section only describes an architecture for encoding a single I/O example. However, in general we assume the input to consist of multiple I/O examples. The number of I/O examples can be variable between test instances, and the examples are unordered, which suggests a pooling-based approach. Previous work \cite{parisotto2017} has pooled on the final encoder hidden states, but this approach cannot be used for attentional models.

Instead, we take an approach which we refer to as \textit{late pooling}. Here, each I/O example has its own layers for $I$, $O$, and $P$ (with shared weights across examples), but the hidden states of $P_1, ..., P_n$ are pooled at each timestep before being fed into a single output softmax layer. The architecture is shown at the bottom of Figure~\ref{fig:synthesis_arch_diagram}. We did not find it beneficial to add another fully-connected layer or recurrent layer after pooling.

Formally, the layers labeled ``FC'' and ``MaxPool'' perform the operation $m_i = {\rm MaxPool}_{j \in n}({\rm tanh}(W{h_{ji}}))$, where $i$ is the current timestep, $n$ is the number of observed examples, $h_{ji} \in \mathbb{R}^d$ is the output of $P_j$ at the timestep $i$, and $W \in \mathbb{R}^{d \times d}$ is a set of learned weights. The layer denoted as ``Output Softmax'' performs the operation $y_i = {\rm Softmax}(Vm_i)$, where $V \in \mathbb{R}^{d \times v}$ is the output weight matrix, and $v$ is the number of tokens in the program vocabulary. The model is trained to maximize the log-softmax of the reference program sequence, as is standard.

\subsection{Hyperparameters and Training}

In all experiments, the size of the recurrent and fully connected layers is 512, and the size of the embeddings is 128. Models were trained with plain SGD + gradient clipping. All models were trained for 2 million minibatch updates, where each minibatch contained 128 training instances (i.e., 128 programs with four I/O examples each). Each minibatch was re-sampled, so the model saw  256 million random programs and 1024 million random I/O examples during training. Training took approximately 24 hours of 2 Titan X GPUs, using an in-house toolkit. A small amount of hyperparameter tuning was done on a synthetic validation set that was generated like the training.

\section{Program Synthesis Results}

Once training is complete, the synthesis models can be decoded with a beam search decoder \cite{sutskever2014}. Unlike a typical sequence generation task, where the model is decoded with a beam $k$ and then only the 1-best output is taken, here all $k$-best candidates are executed one-by-one to determine consistency. If multiple program candidates are consistent with all observed examples, the program with the highest model score is taken as the output.\footnote{We tried several alternative heuristics, such as taking the shortest program, but these did not perform better.} This program is referred to as $P^*$.

In addition to standard beam search, we also propose a variant referred to as ``DP-Beam,'' which adds a search constraint similar to the dynamic programming algorithm mentioned in Section~\ref{sec:data_synthesis}. Here, each time an expression is completed during the search, the partial program is executed in a black-box manner. If any resulting partial {\tt OutStr} is not a string prefix of the observed {\tt OutStr}, the partial program is removed from the beam. This technique is effective because our DSL is largely concatenative.

\begin{figure}[htb]
\begin{centering}
\includegraphics[width=225px]{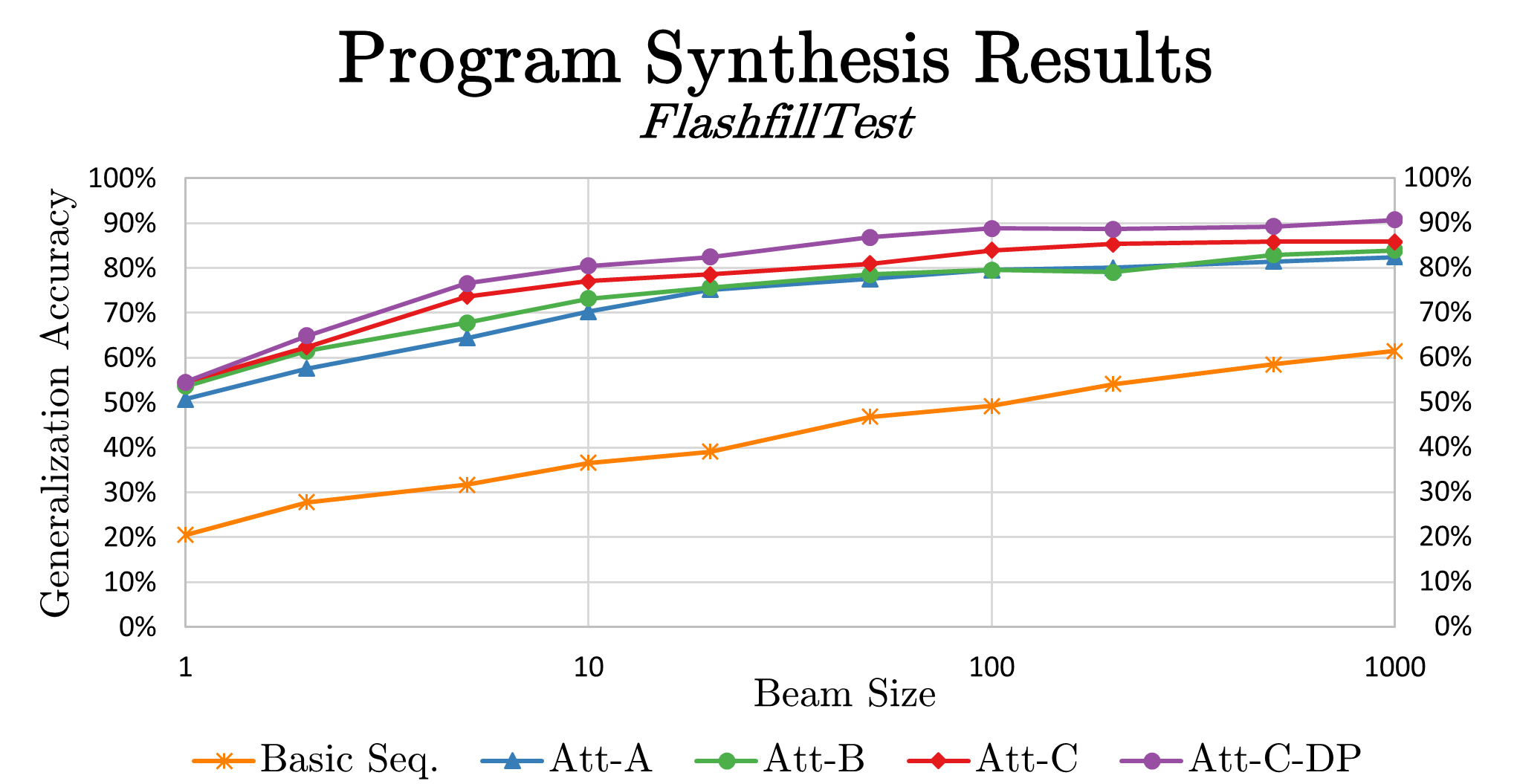}
\caption{Generalization results for program synthesis using several network architectures.}
\label{fig:synthesis_results}
\end{centering}
\end{figure}

Generalization accuracy is computed by applying $P^*$ to all six assessment examples. The percentage score reported in the figures represents the proportion of test instances for which a consistent program was found \textit{and} it resulted in the exact correct output for all six assessment examples. Consistency is evaluated in Section~\ref{sec:con_vs_gen_results}.

Results are shown in Figure~\ref{fig:synthesis_results}. The most evident result is that all attentional variants outperform the basic seq-to-seq model by a very large margin -- roughly 25\% absolute improvement. The difference between the three variants is smaller, but there is a clear improvement in accuracy as the models progress in complexity. Both Attention-B and Attention-C each add roughly 2-5\% absolute accuracy, and this improvement appears even for a large beam. The DP-Beam variant also improves accuracy by roughly 5\%. Overall, the best absolute accuracy achieved is 92\% by Attention-C-DP w/ Beam=1000. Although we have not optimized our decoder for speed, the amortized end-to-end cost of decoding is roughly 0.3 seconds per test instance for Attention-C-DP w/ Beam=100 and four observed examples (89\% accuracy), on a Titan X GPU.

\subsection{Comparison to Past Work}
\label{sec:past_work}

Prior to this work, the strongest statistical model for solving FlashFillTest was \citet{parisotto2017}. The generalization accuracy is shown below:
\begin{center}
\begin{tabular}{|l|c|c|}
\hline
\bf System & \multicolumn{2}{|c|}{\bf Beam} \\ \cline{2-3}
& \bf 100 & \bf 1000 \\ \hline
\citet{parisotto2017} & 23\% & 34\% \\ \hline
Basic Seq-to-Seq & 51\% & 56\% \\ \hline
Attention-C & 83\% & 86\% \\ \hline
Attention-C-DP & 89\% & 92\% \\ \hline
\end{tabular}
\end{center}
We believe that this improvement in accuracy is due to several reasons. First, late pooling allows us to effectively incorporate powerful attention mechanisms into our model. Because the architecture in \citet{parisotto2017} performed pooling at the I/O encoding level, it could not exploit the attention mechanisms which we show our critical to achieving high accuracy. Second, the DSL used here is more expressive, especially the {\tt GetSpan()} function, which was required to solve approximately 20\% of the test instances. \footnote{However, this increased the search space of the DSL by 10x.}

Comparison to the FlashFill implementation currently deployed in Microsoft Excel is given in Section~\ref{sec:noise}.

\subsection{Consistency vs. Generalization Results}
\label{sec:con_vs_gen_results}
\vspace{-10pt}
\begin{figure}[hbt]
\centering
\includegraphics[width=215px]{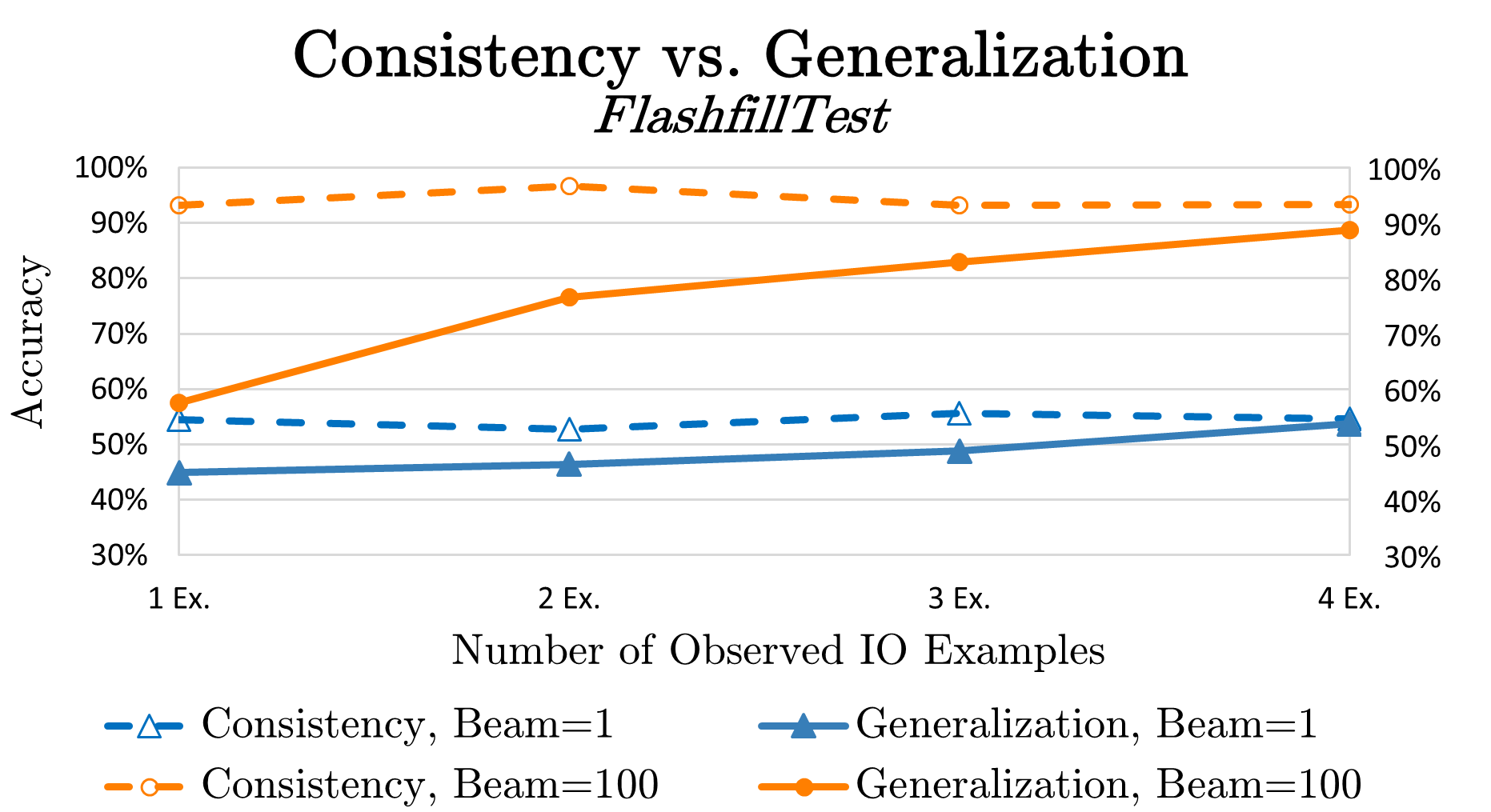}
\caption{Results were obtained using Attention-C.}
\label{fig:con_vs_gen_results}
\end{figure}
\vspace{-10pt}

The conceptual difference between {\it consistency} and {\it generalization} is detailed in Section~\ref{sec:con_vs_gen_desc}. Results for different beam sizes and different number of observed IO examples are presented in Figure~\ref{fig:con_vs_gen_results}. As expected, the generalization accuracy increases with the number of observed examples for both beam sizes, although this is significantly more pronounced for a Beam=100. Interestingly, the consistency is relatively constant when the number of observed examples increases. There was no {\it a priori} expectation about whether consistency would increase or decrease, since more examples are consistent with fewer total programs, but also give the network a stronger input signal. Finally, we can see that the Beam=1 decoding only generates {\it consistent} output roughly 50\% of the time, which implies that the latent function semantics learned by the model are still far from perfect.
 


\section{Program Induction Results}
\label{sec:program_induction}

An alternative approach to solving the FlashFill problem is {\it program induction}, where the output string is generated directly by the neural network without the need for a DSL. More concretely, we can train a neural network which takes as input a set of $n$ observed examples ${ (I_1, O_1), ... (I_n, O_n) }$ as well an unpaired {\tt InStr}, $I^y$, and generates the corresponding {\tt OutStr}, $O^y$. As an example, from Figure~\ref{fig:main_example}, $I_1 = $ ``{\small {\tt john Smith}}'', $O_1 = $ ``{\small {\tt Smith, Jhn}}'', $I_2 = $ ``{\small {\tt DOUG Q. Macklin}}'', ... , $I^y = $ ``{\small {\tt Steve P. Green}}'', $O^y =$ ``{\small {\tt Green, Steve}}''. Both approaches have the same end goal -- determine the $O^y$ corresponding to $I^y$ -- but have several important conceptual differences.

The first major difference is that the induction model does not use the program $P$ anywhere. The synthesis model generates $P$, which is executed by the DSL to produced $O^y$. The induction model generates $O^y$ directly by sequentially predicting each character. In fact, in cases where it is possible to obtain a very large amount of real-world I/O example sets, induction is a very appealing approach since it does not require an explicit DSL.\footnote{In the results shown here, the induction model is trained on data synthesized with the DSL, but the model training is agnostic to this fact.} The core idea is the model learns some {\it latent} program representation which can generalize beyond a specific DSL. It also eliminates the need to hand-design the DSL, unless the DSL is needed to synthesize training data.

The second major difference is that program induction has no concept of {\it consistency}. As described previously, in program synthesis, a $k$-best list of program candidates is executed one-by-one, and the first program consistent with all observed examples is taken as the output.
As shown in Section~\ref{sec:con_vs_gen_results}, if a consistent program can be found, it is likely to generalize to new inputs. 
Program induction, on the other hand, is essentially a standard sequence generation task akin to neural machine translation or image captioning -- we directly decode $O^y$ with a beam search and take the highest-scoring candidate as our output.

\subsection{Comparison of Induction and Synthesis Models}

Despite these differences, it is possible to model both approaches using nearly-identical network architectures. The induction model evaluated here is identical to synthesis Attention-A with late pooling, except for the following two modifications:
\begin{enumerate}
\item Instead of generating $P$, the system generates the new {\tt OutStr} $O^y$ character-by-character.
\item There is an additional LSTM to encode $I^y$. The decoder layer $O^y$ uses double attention on $O_j$ and $I^y$.
\end{enumerate}

The induction network diagram is given in the supplementary material. Each $(I^y, O^y)$ pair is decoded independently, but conditioned on all observed examples. The attention, pooling, hidden sizes, training details, and decoder are otherwise identical to synthesis. The induction model was trained on the same synthetic data as the synthesis models. 

\vspace{-5pt}
\begin{figure}[h]
\centering
\includegraphics[width=225px]{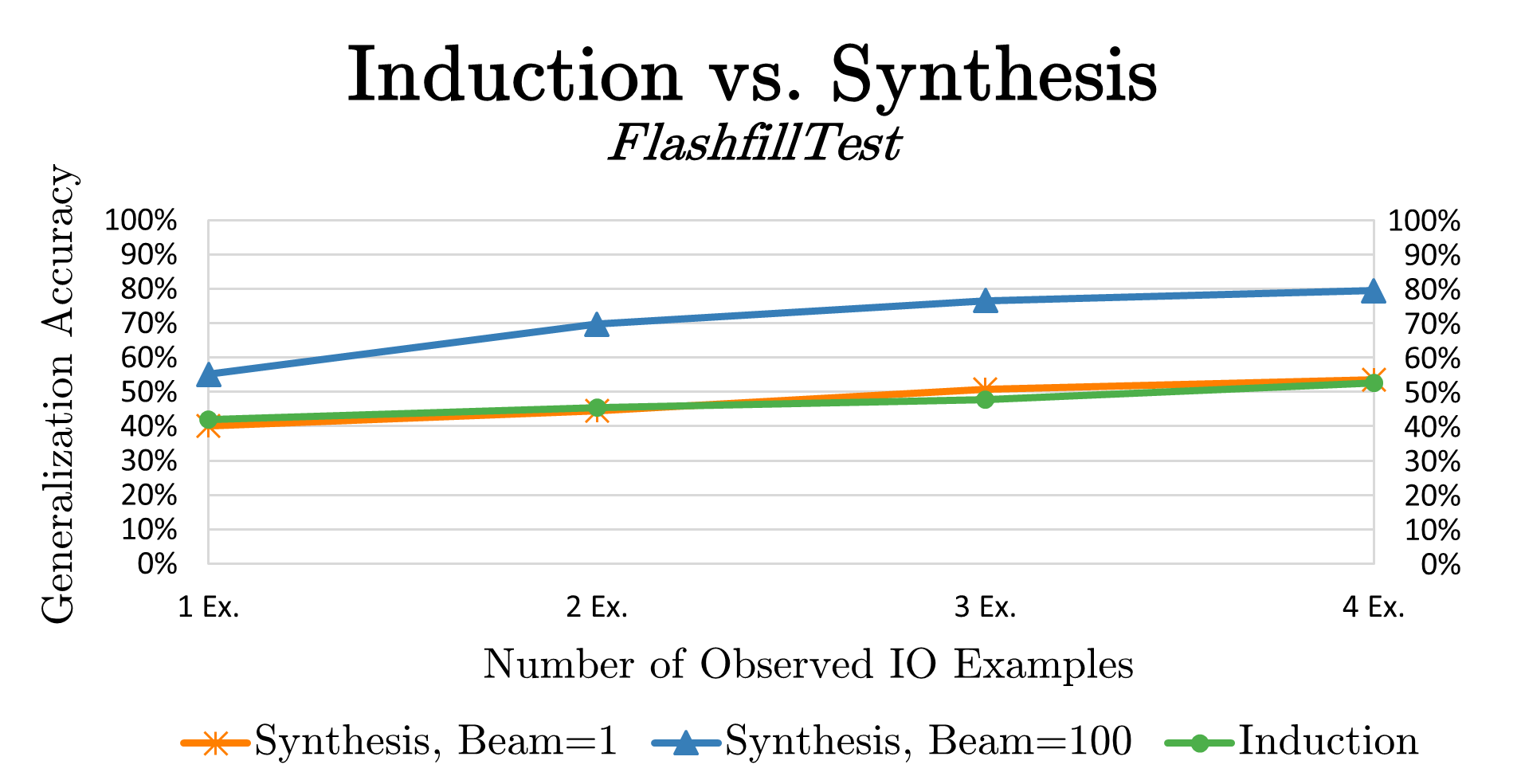}
\caption{The synthesis model uses Attention-A + standard beam search.}
\label{fig:induction_results}
\end{figure}

Results are shown in Figure~\ref{fig:induction_results}. The induction model is compared to synthesis Attention-A using the same measure of generalization accuracy as previous sections -- all six assessment examples must be exactly correct. Induction performs similarly to synthesis w/ beam=1, but both are significantly outperformed by synthesis w/ beam=100. The  generalization accuracy achieved by the induction model is 53\%, compared to 81\% for the synthesis model. The induction model uses a beam of 3, and does not improve with a larger search because there is no way to evaluate candidates after decoding.

\subsection{Average-Example Accuracy}

All previous sections have used a strict definition of ``generalization accuracy,'' requiring all six assessment examples to be exactly correct. We refer to this as {\it all-example} accuracy. However, another useful metric is to measure the total percent of correct assessment examples, averaged over all instances.\footnote{The example still must be exactly correct -- character edit rate is not measured here.} With this metric, generalizing on 5-out-of-6 assessment examples accumulates more credit than 0. We refer to this as {\it average-example} accuracy.

\begin{figure}[ht]
\centering
\includegraphics[width=215px]{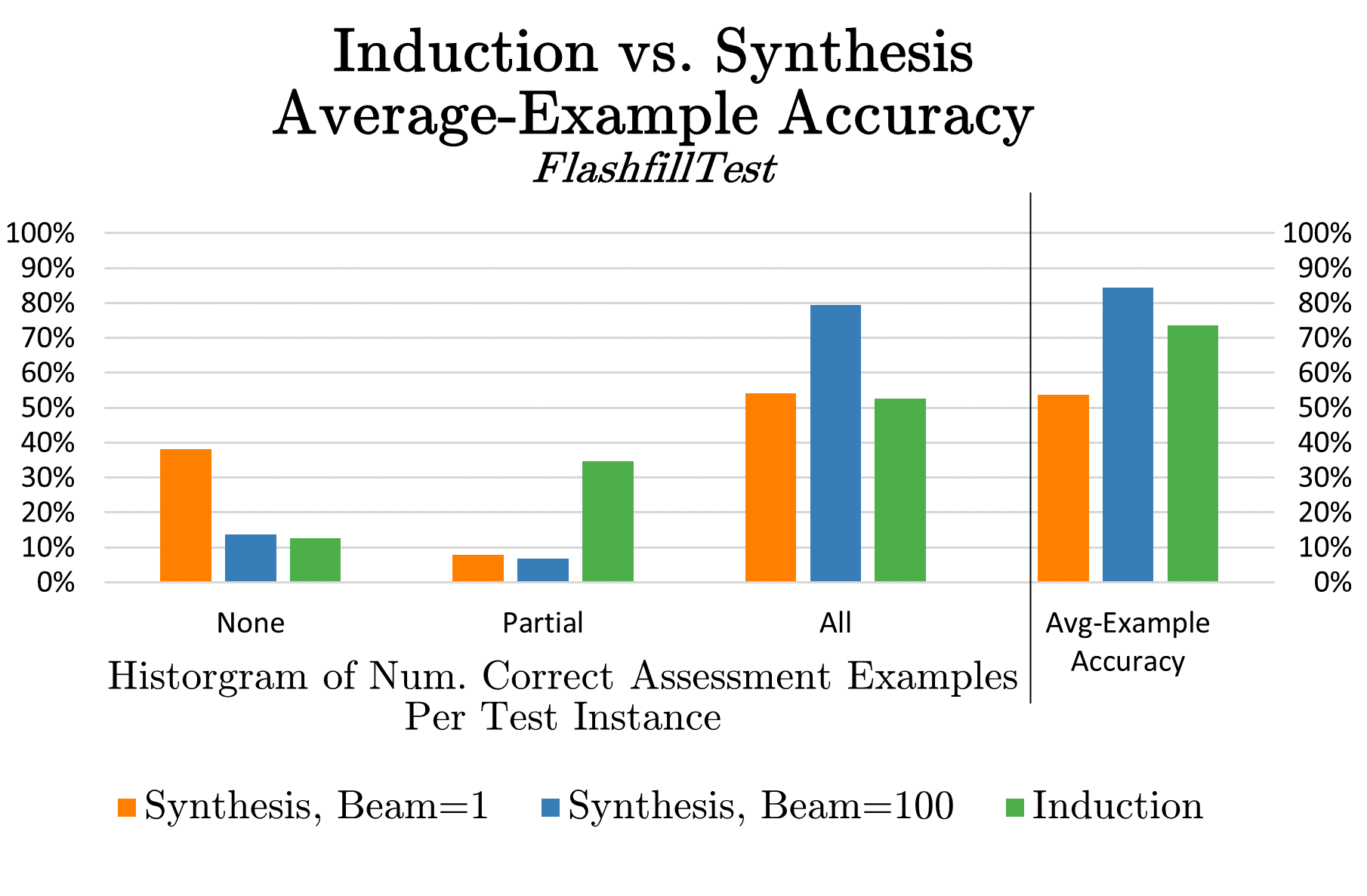}
\vspace{-15pt}
\caption{ All experiments use four observed examples.
}
\vspace{-20pt}
\label{fig:ind_vs_synth_per_example}
\end{figure}

Average-example results are presented in Figure~\ref{fig:ind_vs_synth_per_example}.
The outcome matches our intuitions: Synthesis models tend to be ``all or nothing,'' since it must find a {\it single} program that is jointly consistent with all observed examples. For both synthesis conditions, less than 10\% of the test instances are partially correct. Induction models, on the other hand, have a much higher chance of getting {\it some} of the assessment examples correct, since they are decoded independently. Here, 33\% of the test instances are partially correct. 
Examining the right side of the figure, the induction model shows relative strength under the average-example accuracy metric.
However, in terms of absolute performance, the synthesis model still bests the induction model by 10\%.

It is difficult to suggest which metric should be given more credence, since the utility depends on the downstream application. For example, if a user wanted to automatically fill in an entire column in a spreadsheet, they may prioritize all-example accuracy -- {\it If} the system proposes a solution, they can be confident it will be correct for all rows. However, if the application instead offered auto-complete suggestions on a {\it per-cell} basis, then a model with higher average-example accuracy might be preferred.

\vspace{-5pt}
\section{Handling Noisy I/O Examples}
\label{sec:noise}

For the FlashFill task, real-world I/O examples are typically manually composed by the user, so noise (e.g., typos) is expected and should be well-handled. An example is given in Figure~\ref{fig:main_example}.

Because neural network methods (1) are inherently probabilistic, and (2) operate in a continuous space representation, it is reasonable to believe that they can  learn to be robust to this type of noise. In order to explicitly account for noise, we only made two small modifications. First, noise was synthetically injected into the training data using random character transformations.\footnote{This did not degrade the results on the noise-free test set.} Second, the best program $P^*$ was selected by using {\it character edit rate} (CER) \cite{marzal1993} to the observed examples, rather than exact match.\footnote{Standard beam is also used instead of DP-Beam.}

Since the FlashFillTest set does not contain any noisy examples, noise was synthetically injected into the observed examples. All noise was applied with uniform random probability into the {\tt InStr} or {\tt OutStr} using character insertions, deletions, or substitutions. Noise is not applied to the {\it assessment} examples, as this would make evaluation impossible.

We compare the models in this paper to the actual FlashFill implementation found in Microsoft Excel, as described in \citet{gulwani2012}. An overview of this model is described in Section~\ref{sec:related_work}. The results were obtained using a macro in Microsoft Excel 2016.

\vspace{-10pt}
\begin{figure}[th]
\begin{centering}
\includegraphics[width=215px]{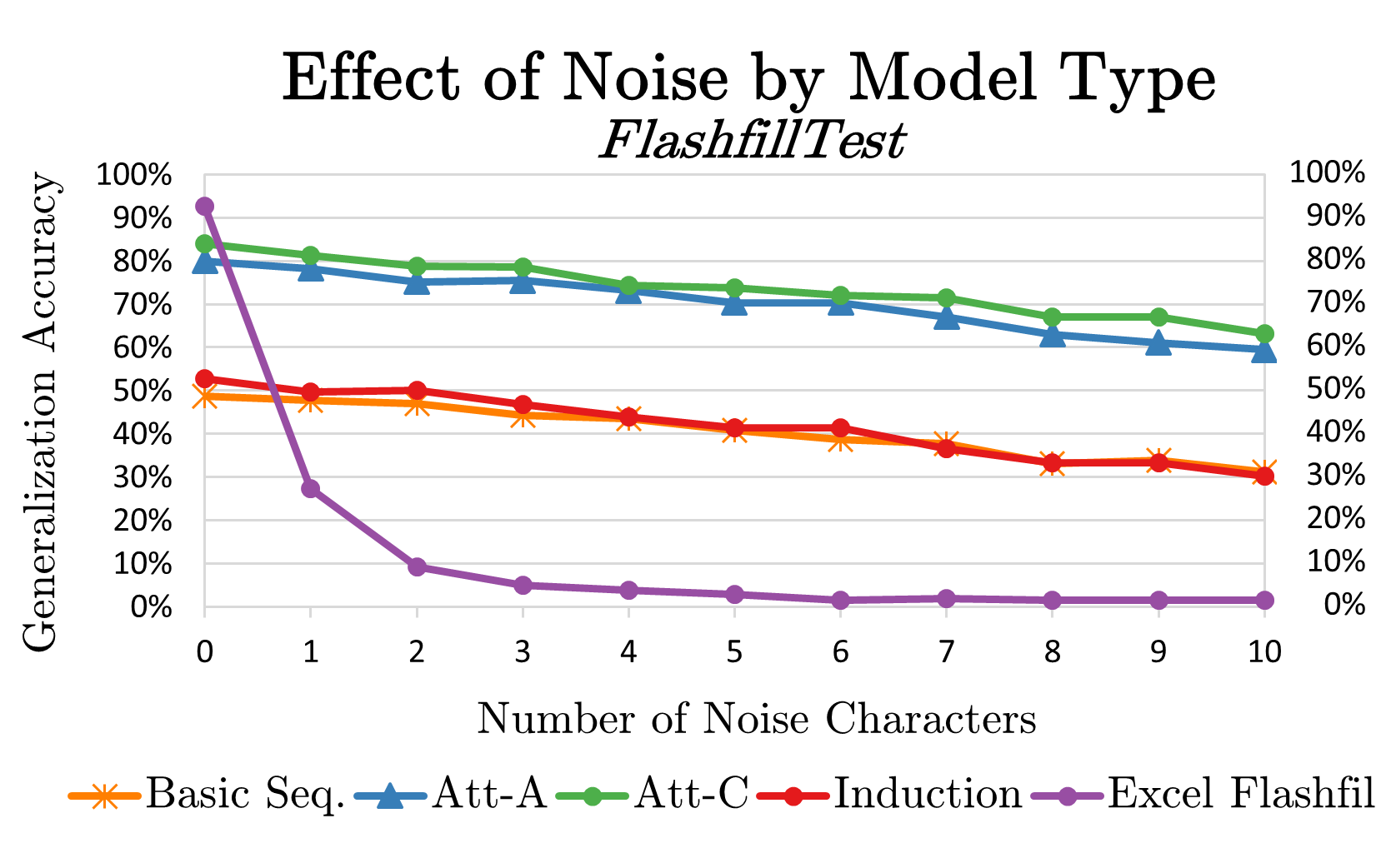}
\vspace{-10pt}
\caption{All results use four observed examples, and all synthesis models use beam=100.}
\label{fig:noise_results}
\end{centering}
\end{figure}
\vspace{-7pt}

The noise results are shown in Figure~\ref{fig:noise_results}. The neural models behave very similarly, each degrading approximately 2\% absolute accuracy for each noise character introduced. The behavior of Excel FlashFill is quite different. Without noise, it achieves 92\% accuracy,\footnote{FlashFill was manually developed on this exact set.} matching the best result reported earlier in this paper. However, with just one or two characters of noise, Excel FlashFill is effectively ``broken.'' This result is expected, since the efficiency of their algorithm is critically centered around exact string matching \cite{gulwani2012}. We believe that this robustness to noise is one of the strongest attributes of DNN-based approaches to program synthesis.

\vspace{-5pt}
\section{Conclusions}
\label{sec:conclusions}
We have presented a novel variant of an attentional RNN architecture for program synthesis which achieves 92\% accuracy on a real-world Programming By Example task. This matches the performance of a hand-engineered system and outperforms the previous-best neural synthesis model by 58\%. Moreover, we have demonstrated that our model remains robust to moderate levels of noise in the I/O examples, while the hand-engineered system fails for even small amounts of noise. Additionally, we carefully contrasted our neural program synthesis system with a neural program induction system, and showed that even though the synthesis system performs better on this task, both approaches have their own strength under certain evaluation conditions. In particular, synthesis systems have an advantage when evaluating if {\it all} outputs are correct, while induction systems have strength when evaluating which system has the {\it most} correct outputs.

\bibliography{ms}
\bibliographystyle{icml2017}

\onecolumn

\begin{center}\Huge
{\bf Supplementary Material}
\end{center}

\appendix 

\section{DSL Extended Description}

\renewcommand{\t}[1]{\texttt{#1}}
\newcommand{\sem}[1]{\llbracket #1 \rrbracket_{v}}
\newcommand{\sems}[1]{\llbracket #1 \rrbracket_{v}}

Section 3.2 of the paper provides the grammar of our domain specific language, which both defines the space of possible programs, and allows us to easily sample programs. The formal semantics of this language are defined below in Figure~\ref{fig:dslsemantics}. The program takes as input a string $v$ and produces a string as output (result of \t{Concat} operator).

As an implementational detail, we note that after sampling a program from the grammar, we flatten calls to nesting functions (as defined in Figure 2 of the paper) into a single token. For example, the function {\tt GetToken(t, i)} would be tokenized as a single token ${\tt GetToken_{t,i}}$ rather than $3$ separate tokens. This is possible because for nesting functions, the size of the total parameter space is small. For all other functions, the parameter space is too large for us to flatten function calls without dramatically increasing the vocabulary size, so we treat parameters as separate tokens.

\vspace{30pt}

\begin{figure}[H]
\centering
\shiftleft{225pt}{ 
\scalebox{0.8}{\parbox{\linewidth}{
\begin{eqnarray*}
\sem{\t{Concat}(e_1,e_2,e_3, ...)} & = & \t{Concat}(\sem{e_1}, \sem{e_2}, \sem{e_3}, ...) \\
\sem{n_1(n_2)} & = & \llbracket n_1 \rrbracket_{v_1}, \mbox{ where } v_1 = \sem{n_2}\\
\sem{n(f)} & = & \llbracket n \rrbracket_{v_1}, \mbox{ where } v_1 = \sem{f}\\
\sem{\t{ConstStr}(c)} & = & c \\
\sem{\t{SubStr}(k_1,k_2)} & = & v[p_1 .. p_2], \mbox{ where}\\
& & p_1 = k_1 > 0 \; ? \; k_1 \; : \; \t{len(v)} + k_1 \\
& & p_2 = k_2 > 0 \; ? \; k_2 \; : \; \t{len(v)} + k_2 \\
\sem{\t{GetSpan}(r_1,i_1,y_1,r_2,i_2,y_2)} & = & v[p_1 .. p_2] \mbox{ ,where}\\
& & p_1 = y_1 \mbox{(Start or End) of } |i_1|^\t{th} \mbox{ match of } r_1 \mbox{ in } v \mbox{ from beginning (end if } i_i < 0 \mbox{)}\\
& & p_2 = y_2 \mbox{(Start or End) of } |i_2|^\t{th} \mbox{ match of } r_2 \mbox{ in } v \mbox{ from beginning (end if } i_2 < 0 \mbox{)}\\
\sem{\tt{GetToken}(t, i) } & = & |i|^\t{th} \mbox{ match of } t \mbox{ in } v \mbox{ from beginning (end if } i<0 \mbox{)}\\
\sem{\tt{GetUpto}(r) } & = & v [0..i], \mbox{ where $i$ is the index of end of first match of } r \mbox{ in } v \mbox{ from beginning}\\
\sem{\tt{GetFrom}(r) } & = & v [j..len(v)], \mbox{ where j is the end of last match of } r \mbox{ in } v \mbox{ from end}\\
\sem{\tt{GetFirst}(t, i) } & = & \t{Concat}(s_1,\cdots,s_i), \mbox{ where $s_j$ denotes the $j^\t{th}$ match of } t \mbox{ in } v\\
\sem{\tt{GetAll}(t) } & = & \t{Concat}(s_1,\cdots,s_m), \mbox{ where $s_i$ denotes the $i^\t{th}$ match of } t \mbox{ in } v \mbox{ and $m$ denotes the total matches}\\
\sem{\tt{ToCase}(s)} &=& \t{ToCase}(s,v) \\
\sem{\tt{Trim}()} &=& \t{Trim}(v) \\
\sem{\tt{Replace}(\delta_1,\delta_2)} & = & \t{Replace}(v, \delta_1,\delta_2) \\
\end{eqnarray*}
}}
}
\caption{ The semantics of the DSL for string transformations.}
\label{fig:dslsemantics}
\end{figure}

\section{Synthetic Evaluation Details}

Results on synthetically generated examples are largely omitted from the paper since, in a vacuum, the synthetic dataset can be made arbitrarily easy or difficult via different generation procedures, making summary statistics difficult to interpret. We instead report results on an external real-world dataset to verify that the model has learned function semantics which are at least as expressive as programs observed in real data.

Nevertheless, we include additional details about our experiments on synthetically generated programs for readers interested in the details of our approach. As described in the paper, programs were randomly generated from the DSL by first determining a program length up to a maximum of 10 \textit{expressions}, and then independently sampling each expression. We used a simple set of heuristics to restrict potential inputs to strings which will produce non-empty outputs (e.g. any program which references the third occurrence of a number will cause us to sample strings containing at least three numbers). We rejected any degenerate samples e.g. those resulting in empty outputs, or outputs longer than 100 characters.

Figure~\ref{synth_samples} shows several random synthetically generated samples.

\begin{figure}[H]
\centering
\includegraphics[width=375px]{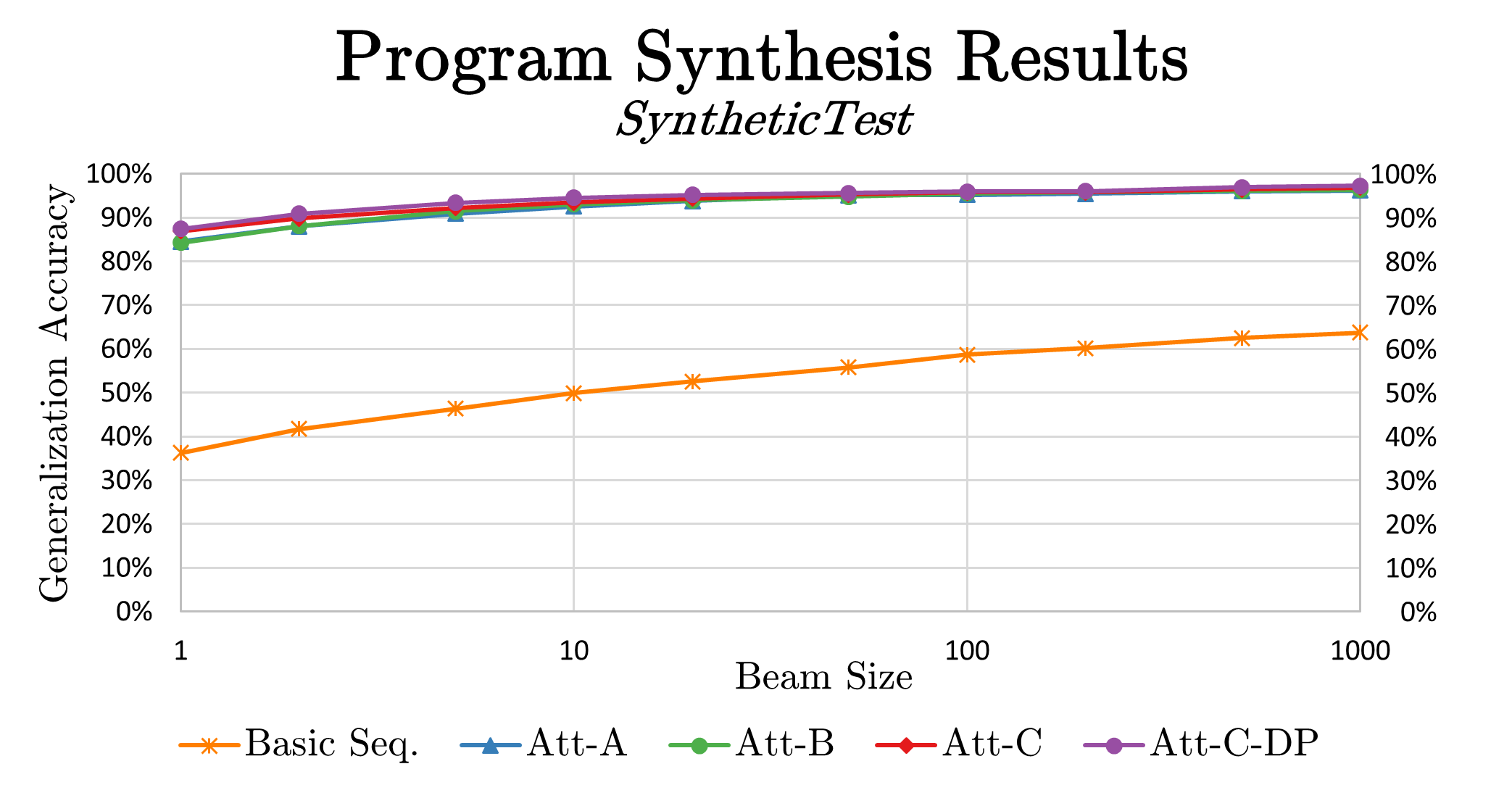}
\caption{Generalization accuracy for different models on the synthetic validation set}
\label{fig:synthetic_results}
\end{figure}

Figure~\ref{fig:synthetic_results} shows the accuracy of each model on the synthetically generated validation set. Model accuracy on the synthetic validation set is generally consistent with accuracy on the FlashFill dataset, with stronger models on the synthetic dataset also demonstrating stronger performance on the real-world data.

\vspace{30pt}

\section{Examples of Synthesized Programs}

Figure~\ref{fig:random_ff_decoding} shows several randomly sampled (anonymized) examples from the FlashFill test set, along with their predicted programs outputted by the synthesis model.

Figure~\ref{fig:consistent_samples} shows several examples which were hand-selected to demonstrate interesting limitations of the model. In the case of the first example, the task is to reformat international telephone numbers. Here, the task is underconstrained given the observed input-output examples, because there are many different programs which are consistent with the observed examples. Note that to extract the first two digits, there are many other possible functions which would produce the correct output in the observed examples, some of which would generalize and some which would not: for exampling, getting the second and third characters, getting the first two digits, or getting the first number. In this case, the predicted program extracts the country code by taking the first two digits, a strategy which fails to generalize to examples with different country codes.
The third example demonstrates a difficulty of using real world data. Because examples can come from a variety of sources, they may be irregularly formatted. In this case, although the program is consistent with the observed examples, it does not generalize when the second space in the address is removed. In the final example, the synthesis model completely fails, and none of the 100 highest scoring programs from the model were consistent with the observed output examples. The selected program is the closest program scored by character edit distance.

\section{Induction Network Architecture}

The network architecture used in the program induction setting is described in Section 6.1 of the paper. The network structure is a modification of synthesis Attention-A, using double attention to jointly attend to $I^x$ and $O_j$, and an additional LSTM to encode $I^x$.
We include a complete diagram below in Figure~\ref{fig:induction_arch}.

\vspace{30pt}

\begin{figure}[hbt]
\centering
\includegraphics[width=225px]{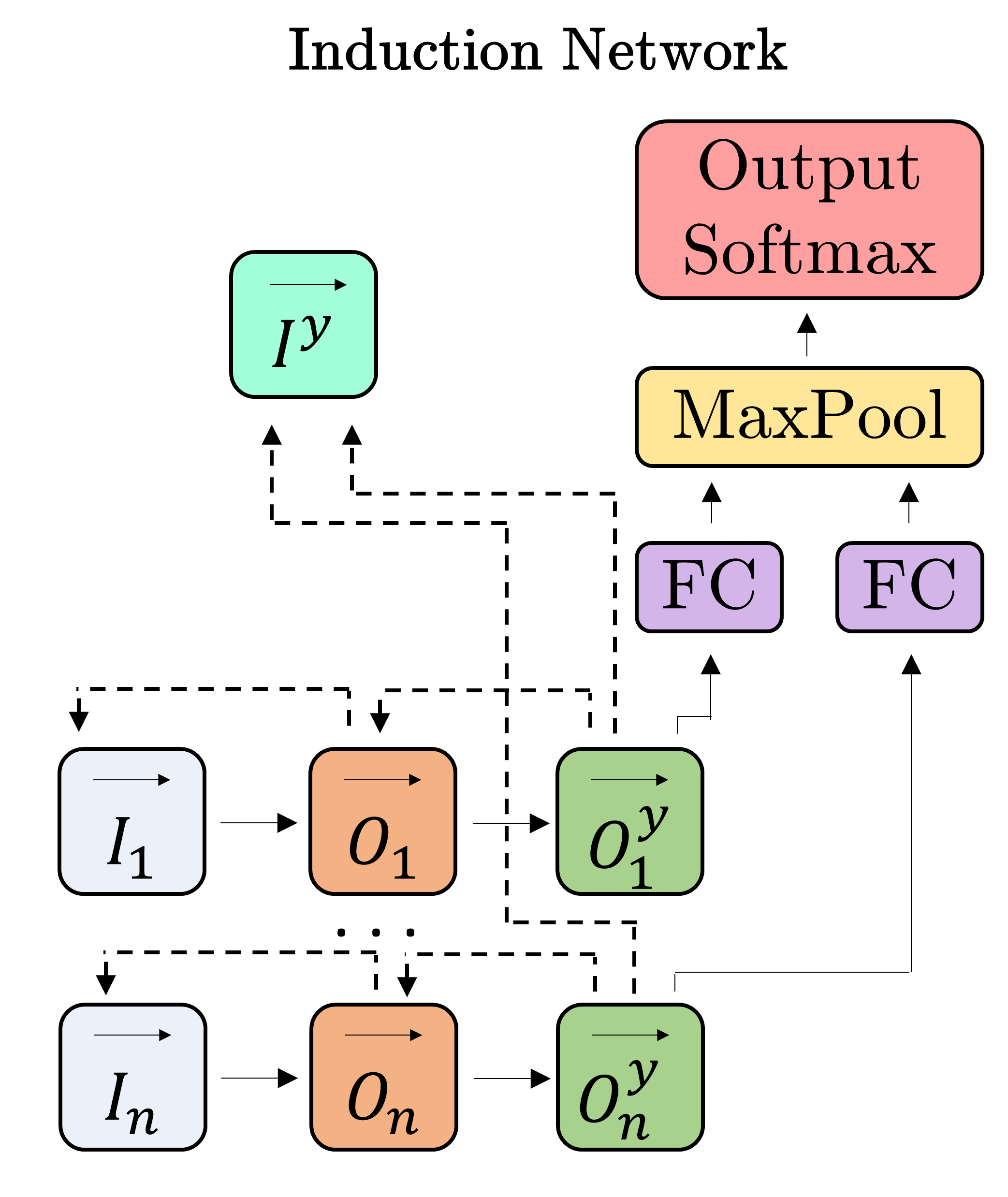}
\caption{The network architecture used for program induction. A dotted line from $x$ to $y$ means that $x$ attends to $y$.}
\label{fig:induction_arch}
\end{figure}

\onecolumn
\begin{figure}

{\centering 
\shiftleft{-10pt}{ 
\begin{tabular}{|p{0.55\linewidth}|p{0.35\linewidth}|}
\hline
\multicolumn{2}{ |p{0.9\linewidth}| }{{\tt Reference program:  GetToken{\uscore}Alphanum{\uscore}3 | GetFrom{\uscore}Colon | GetFirst{\uscore}Char{\uscore}4 }} \\ \hline
{\tt Ud 9:25,JV3 Obb } & {\tt 2525,JV3 ObbUd92 } \\
{\tt zLny xmHg 8:43 A44q } & {\tt 843 A44qzLny } \\
{\tt A6 g45P 10:63 Jf } & {\tt 1063 JfA6g4 } \\
{\tt cuL.zF.dDX,12:31 } & {\tt dDX31cuLz } \\
{\tt ZiG OE bj3u 7:11 } & {\tt bj3u11ZiGO } \\
\hline
\end{tabular}
}}

\vspace{0.6cm}


{\centering 
\shiftleft{-10pt}{ 
\begin{tabular}{|p{0.55\linewidth}|p{0.35\linewidth}|}
\hline
\multicolumn{2}{ |p{0.9\linewidth}| }{\tt
 Reference program: Get{\uscore}Word{\uscore}-1(GetSpan(Word, 1, Start, `(', 5, Start)) | GetToken{\uscore}Number{\uscore}-5 | GetAll{\uscore}Proper | SubStr(-24, -14) | GetToken{\uscore}Alphanum{\uscore}-2 | EOS  } \\ \hline
{\tt 4 Kw ( )SrK (11 (3 CHA xVf )4 )8 Qagimg ) ( )(vs } & {\tt Qagimg4Kw Sr Vf QagimgVf )4 )8 QaQagimg } \\ \hline
{\tt iY) )hspA.5 ( )8,ZsLL (nZk.6 (E4w )2(Hpprsqr )2(Z } & {\tt Hpgjprsqr8Zs Zk Hpprsqrk.6 (E4w )22 } \\ \hline
{\tt Cqg) ) ( (1005 ( ( )VCE hz ) (10 Hadj )zg Tqwpaxft-7 5 6 } & {\tt hz10005Cqg Hadj Tqwpaxft Hadj )zg T5 } \\ \hline
{\tt JvY) (Ihitux ) ) ( (6 SFl (7 XLTD sfs ) )11,lU7 (6 9 } & {\tt lU7Jv Ihitux Frl XLTD sfs )6 } \\ \hline
{\tt NjtT(D7QV (4 (yPuY )8.sa ( ) )6 aX 4 )DXR ( @6 ) Ztje } & {\tt DXR4Njt Pu Ztje)6 aX 4 )DX6 } \\ \hline
\end{tabular}
}}

\vspace{0.6cm}


{\centering 
\shiftleft{-10pt}{ 
\begin{tabular}{|p{0.55\linewidth}|p{0.35\linewidth}|}
\hline
\multicolumn{2}{ |p{0.9\linewidth}| }{\tt Reference program: GetToken{\uscore}AllCaps{\uscore}-2(GetSpan(AllCaps, 1, Start, AllCaps, 5, Start)) | EOS} \\ \hline
{\tt YDXJZ @ZYUD Wc-YKT GTIL BNX } & {\tt W } \\
{\tt JUGRB.MPKA.MTHV,tEczT-GZJ.MFT } & {\tt MTHV } \\
{\tt VXO.OMQDK.JC-OAR,HZGH-DJKC } & {\tt JC } \\
{\tt HCUD-WDOC,RTTRQ-KVETK-whx-DIKDI } & {\tt RTTRQ } \\
{\tt JFNB.Avj,ODZBT-XHV,KYB @,RHVVW } & {\tt ODZBT } \\
\hline
\end{tabular}
}}

\vspace{0.6cm}

{\centering 
\shiftleft{-10pt}{ 
\begin{tabular}{|p{0.55\linewidth}|p{0.35\linewidth}|}
\hline
\multicolumn{2}{ |p{0.9\linewidth}| }{\tt Reference program: SubStr(-20, -8) | GetToken{\uscore}AllCaps{\uscore}-3 | SubStr(11, 19) | GetToken{\uscore}Alphanum{\uscore}-5 | EOS } \\ \hline
{\tt DvD 6X xkd6 OZQIN ZZUK,nCF aQR IOHR } & {\tt IN ZZUK,nCF aCFv OZQIN ZOZQIN } \\ \hline
{\tt BHP-euSZ,yy,44-CRCUC,ONFZA.mgOJ.Hwm } & {\tt CRCUC,ONFZA.mONFZAy,44-CRCU44 } \\ \hline
{\tt NGM-8nay,xrL.GmOc.PFLH,CMFEX-JPFA,iIcj,329 } & {\tt ,CMFEX-JPFA,iCMFEXrL.GmOc.PPFLH } \\ \hline
{\tt hU TQFLD Lycb NCPYJ oo FS TUM l6F } & {\tt NCPSYJ oo FS FScb NCPYJ NCPYJ } \\ \hline
{\tt OHHS NNDQ XKQRN KDL 8Ucj dUqh Cpk Kafj } & {\tt L 8Ucj dUqh CUXKQRN KDLKDL } \\
\hline
\end{tabular}
}}

\vspace*{0.6cm}
\caption{Randomly sampled programs and corresponding input-output examples, drawn from training data. Multi-line examples are all broken into lines on spaces.}
\label{synth_samples}
\end{figure}

\begin{figure}
\subfloat
{\centering 
\shiftleft{-10pt}{ 
\begin{tabular}{|p{0.45\linewidth}|p{0.225\linewidth}|p{0.225\linewidth}|}
\hline
\multicolumn{3}{ |p{0.9\textwidth}| }
{\texttt{Model prediction: GetSpan(`[', 1, Start, Number, 1, End) | Const(]) | EOS
} } \\
\hline
{\tt [CPT-101 } & {\tt [CPT-101] }  & \textcolor{darkgreen} {\tt [CPT-101] } \\
{\tt [CPT-101 } & {\tt [CPT-101] }  & \textcolor{darkgreen} {\tt [CPT-101] } \\
{\tt [CPT-11] } & {\tt [CPT-11] }  & \textcolor{darkgreen} {\tt [CPT-11] } \\
{\tt [CPT-1011] } & {\tt [CPT-1011] }  & \textcolor{darkgreen} {\tt [CPT-1011] } \\
\hline
{\tt [CPT-1011 } & {\tt [CPT-1011] }  & \textcolor{darkgreen} {\tt [CPT-1011] } \\
{\tt [CPT-1012 } & {\tt [CPT-1012] }  & \textcolor{darkgreen} {\tt [CPT-1012] } \\
{\tt [CPT-101] } & {\tt [CPT-101] }  & \textcolor{darkgreen} {\tt [CPT-101] } \\
{\tt [CPT-111] } & {\tt [CPT-111] }  & \textcolor{darkgreen} {\tt [CPT-111] } \\
{\tt [CPT-1011] } & {\tt [CPT-1011] }  & \textcolor{darkgreen} {\tt [CPT-1011] } \\
{\tt [CPT-101] } & {\tt [CPT-101] }  & \textcolor{darkgreen} {\tt [CPT-101] } \\
\hline
\end{tabular}
}}

\vspace{0.6cm}

\subfloat
{\centering 
\shiftleft{-10pt}{ 
\begin{tabular}{|p{0.25\linewidth}|p{0.325\linewidth}|p{0.325\linewidth}|}
\hline
\multicolumn{3}{ |p{0.9\textwidth}| }
{\texttt{Model prediction: Replace{\uscore}Space{\uscore}Comma(GetSpan(Proper, 1, Start, Proper, 4, End) | Const(.) | GetToken{\uscore}Proper{\uscore}-1 | EOS}} \\
\hline
{\tt Jacob Ethan James Alexander Michael } & {\tt Jacob,Ethan,James,Alexander.{\hyphen}Michael }  & \textcolor{darkgreen} {\tt Jacob,Ethan,James,Alexander.{\hyphen}Michael } \\ \hdashline
{\tt Elijah Daniel Aiden Matthew Lucas } & {\tt Elijah,Daniel,Aiden,Matthew.{\hyphen}Lucas }  & \textcolor{darkgreen} {\tt Elijah,Daniel,Aiden,Matthew.{\hyphen}Lucas } \\ \hdashline
{\tt Jackson Oliver Jayden Chris Kevin } & {\tt Jackson,Oliver,Jayden,Chris.{\hyphen}Kevin }  & \textcolor{darkgreen} {\tt Jackson,Oliver,Jayden,Chris.{\hyphen}Kevin } \\ \hdashline
{\tt Earth Fire Wind Water Sun } & {\tt Earth,Fire,Wind,Water.Sun }  & \textcolor{darkgreen} {\tt Earth,Fire,Wind,Water.Sun } \\
\hline
{\tt Tom Mickey Minnie Donald Daffy } & {\tt Tom,Mickey,Minnie,Donald.Daffy }  & \textcolor{darkgreen} {\tt Tom,Mickey,Minnie,Donald.Daffy } \\ \hdashline
{\tt Jacob Mickey Minnie Donald Daffy } & {\tt Jacob,Mickey,Minnie,Donald.{\hyphen}Daffy }  & \textcolor{darkgreen} {\tt Jacob,Mickey,Minnie,Donald.{\hyphen}Daffy } \\ \hdashline
{\tt Gabriel Ethan James Alexander Michael } & {\tt Gabriel,Ethan,James,Alexander{\hyphen}.Michael }  & \textcolor{darkgreen} {\tt Gabriel,Ethan,James,Alexander.{\hyphen}Michael } \\ \hdashline
{\tt Rahul Daniel Aiden Matthew Lucas } & {\tt Rahul,Daniel,Aiden,Matthew.{\hyphen}Lucas }  & \textcolor{darkgreen} {\tt Rahul,Daniel,Aiden,Matthew.{\hyphen}Lucas } \\ \hdashline
{\tt Steph Oliver Jayden Chris Kevin } & {\tt Steph,Oliver,Jayden,Chris.Kevin }  & \textcolor{darkgreen} {\tt Steph,Oliver,Jayden,Chris.Kevin } \\ \hdashline
{\tt Pluto Fire Wind Water Sun } & {\tt Pluto,Fire,Wind,Water.Sun }  & \textcolor{darkgreen} {\tt Pluto,Fire,Wind,Water.Sun } \\
\hline
\end{tabular}
}}

\vspace{0.6cm}
\subfloat
{\centering 
\shiftleft{-10pt}{ 
\begin{tabular}{|p{0.45\linewidth}|p{0.225\linewidth}|p{0.225\linewidth}|}
\hline
\multicolumn{3}{ |p{0.9\textwidth}| }
{\texttt{Model prediction: GetAll{\uscore}Proper | EOS} } \\
\hline
{\tt Emma Anders } & {\tt Emma Anders }  & \textcolor{darkgreen} {\tt Emma Anders } \\
{\tt Olivia Berglun } & {\tt Olivia Berglun }  & \textcolor{darkgreen} {\tt Olivia Berglun } \\
{\tt Madison Ashworth } & {\tt Madison Ashworth }  & \textcolor{darkgreen} {\tt Madison Ashworth } \\
{\tt Ava Truillo } & {\tt Ava Truillo }  & \textcolor{darkgreen} {\tt Ava Truillo } \\
\hline
{\tt Isabella } & {\tt Isabella }  & \textcolor{darkgreen} {\tt Isabella } \\
{\tt Mia } & {\tt Mia }  & \textcolor{darkgreen} {\tt Mia } \\
{\tt Emma Stevens } & {\tt Emma Stevens }  & \textcolor{darkgreen} {\tt Emma Stevens } \\
{\tt Chris Charles } & {\tt Chris Charles }  & \textcolor{darkgreen} {\tt Chris Charles } \\
{\tt Liam Lewis } & {\tt Liam Lewis }  & \textcolor{darkgreen} {\tt Liam Lewis } \\
{\tt Abigail Jones } & {\tt Abigail Jones }  & \textcolor{darkgreen} {\tt Abigail Jones } \\
\hline
\end{tabular}
}}

\vspace{0.4cm}
\caption{Random samples from the FlashFill test set. The first two columns are {\tt InStr} and {\tt OutStr} respectively, and the third column is the execution result of the predicted program. Example strings which do not fit on a single line are broken on spaces, or hyphenated when necessary. All line-ending hyphens are inserted for readability, and are not part of the example.}
\label{fig:random_ff_decoding}
\end{figure}

\begin{figure}
\ContinuedFloat

\subfloat
{\centering 
\shiftleft{-10pt}{ 
\begin{tabular}{|p{0.45\linewidth}|p{0.225\linewidth}|p{0.225\linewidth}|}
\hline
\multicolumn{3}{ |p{0.9\textwidth}| }
{\texttt{Model prediction: GetToken{\uscore}Proper{\uscore}1 | Const(.) |}} \\
\multicolumn{3}{ |p{0.9\textwidth}| }
{\texttt{GetToken{\uscore}Char{\uscore}1(GetToken{\uscore}Proper{\uscore}-1) | Const(@) | EOS} } \\
\hline
{\tt Mason Smith } & {\tt Mason.S@ }  & \textcolor{darkgreen} {\tt Mason.S@ } \\
{\tt Lucas Janckle } & {\tt Lucas.J@ }  & \textcolor{darkgreen} {\tt Lucas.J@ } \\
{\tt Emily Jacobnette } & {\tt Emily.B@ }  & \textcolor{darkgreen} {\tt Emily.B@ } \\
{\tt Charlotte Ford } & {\tt Charlotte.F@ }  & \textcolor{darkgreen} {\tt Charlotte.F@ } \\
\hline
{\tt Harper Underwood } & {\tt Harper.U@ }  & \textcolor{darkgreen} {\tt Harper.U@ } \\
{\tt Emma Stevens } & {\tt Emma.S@ }  & \textcolor{darkgreen} {\tt Emma.S@ } \\
{\tt Chris Charles } & {\tt Chris.C@ }  & \textcolor{darkgreen} {\tt Chris.C@ } \\
{\tt Liam Lewis } & {\tt Liam.L@ }  & \textcolor{darkgreen} {\tt Liam.L@ } \\
{\tt Olivia Berglun } & {\tt Olivia.B@ }  & \textcolor{darkgreen} {\tt Olivia.B@ } \\
{\tt Abigail Jones } & {\tt Abigail.J@ }  & \textcolor{darkgreen} {\tt Abigail.J@ } \\
\hline
\end{tabular}
}
}

\vspace{0.6cm}

\caption{Random samples from the FlashFill test set. The first two columns are {\tt InStr} and {\tt OutStr} respectively, and the third column is the execution result of the predicted program. Example strings which do not fit on a single line are broken on spaces, or hyphenated when necessary. All line-ending hyphens are inserted for readability, and are not part of the example.}
\label{fig:consistent_samples}
\end{figure}


\begin{figure}

\subfloat
{\centering 
\shiftleft{-10pt}{ 
\begin{tabular}{|p{0.45\linewidth}|p{0.225\linewidth}|p{0.225\linewidth}|}
\hline
\multicolumn{3}{ |p{0.9\textwidth}| }
{\texttt{Model prediction: GetFirst\char`_Digit\char`_2 | Const(.) | GetToken\char`_Number\char`_2 | Const(.) | GetToken{\uscore}Number{\uscore}3 | Const(.) | GetToken\char`_Alpha\char`_-1 | EOS } } \\
\hline
{\tt +32-2-704-33 } & {\tt 32.2.704.33 }  & \textcolor{darkgreen} {\tt 32.2.704.33 } \\
{\tt +44-118-909-3574 } & {\tt 44.118.909.3574 }  & \textcolor{darkgreen} {\tt 44.118.909.3574 } \\
{\tt +90-212-326 5264 } & {\tt 90.212.326.5264 }  & \textcolor{darkgreen} {\tt 90.212.326.5264 } \\
{\tt +44 118 909 3843 } & {\tt 44.118.909.3843 }  & \textcolor{darkgreen} {\tt 44.118.909.3843 } \\ \hline
{\tt +386 1 5800 839 } & {\tt 386.1.5800.839 }  & \textcolor{red} {\tt 38.1.5800.839 } \\
{\tt +1 617 225 2121 } & {\tt 1.617.225.2121 }  & \textcolor{red} {\tt 16.617.225.2121 } \\
{\tt +91-2-704-33 } & {\tt 91.2.704.33 }  & \textcolor{darkgreen} {\tt 91.2.704.33 } \\
{\tt +44-101-909-3574 } & {\tt 44.101.909.3574 }  & \textcolor{darkgreen} {\tt 44.101.909.3574 } \\
{\tt +90-212-326 2586 } & {\tt 90.212.326.2586 }  & \textcolor{darkgreen} {\tt 90.212.326.2586 } \\
{\tt +44 118 212 3843 } & {\tt 44.118.212.3843 }  & \textcolor{darkgreen} {\tt 44.118.212.3843 } \\ \hline
\end{tabular}
}}

\vspace{0.8cm}
\subfloat
{\centering 
\shiftleft{-10pt}{ 
\begin{tabular}{|p{0.45\linewidth}|p{0.225\linewidth}|p{0.225\linewidth}|}
\hline
\multicolumn{3}{ |p{0.9\textwidth}| }{\texttt{ Model prediction: GetFirst\char`_Char\char`_1 | Const(.) | GetFirst\char`_Char\char`_1( GetToken\char`_Proper\char`_4 ) | Const(.) | EOS }} \\ \hline
{\tt Milk 4, Yoghurt 12, Juice 2 Lassi 5 } & {\tt M.L. }  & \textcolor{darkgreen} {\tt M.L. } \\
{\tt Alpha 10 Beta 20 Charlie 40 60 Epsilon } & {\tt A.E. }  & \textcolor{darkgreen} {\tt A.E. } \\
{\tt Sumit 7 Rico 12 Wolfram 15 Rick 19 } & {\tt S.R. }  & \textcolor{darkgreen} {\tt S.R. } \\
{\tt Us 38 China 35 Russia 27 India 1 } & {\tt U.I. }  & \textcolor{darkgreen} {\tt U.I. } \\
\hline
{\tt 10 Apple 2 Oranges 13 Bananas 40 Pears } & {\tt A.P. }  & \textcolor{red} {\tt 1.P. } \\
{\tt 10 Bpple 2 Oranges 13 Bananas 40 Pears } & {\tt B.P. }  & \textcolor{red} {\tt 1.P. } \\
{\tt Milk 4, Yoghurt 12, Juice 2 Massi 5 } & {\tt M.M. }  & \textcolor{darkgreen} {\tt M.M. } \\
{\tt Alpha 10 Beta 20 Charlie 40 60 Delta } & {\tt A.D. }  & \textcolor{darkgreen} {\tt A.D. } \\
{\tt Parul 7 Rico 12 Wolfram 15 Rick 19 } & {\tt P.R. }  & \textcolor{darkgreen} {\tt P.R. } \\
{\tt Us 38 China 35 Russia 27 America 1 } & {\tt U.A. }  & \textcolor{darkgreen} {\tt U.A. } \\
\hline
\end{tabular}
}}

\vspace{0.8cm}
\subfloat
{\centering 
\shiftleft{-10pt}{ 
\begin{tabular}{|p{0.45\linewidth}|p{0.225\linewidth}|p{0.225\linewidth}|}
\hline
\multicolumn{3}{ |p{0.9\textwidth}| } {\texttt{Model prediction: Replace\char`_Space\char`_Dash(GetSpan(AlphaNum, 1, Start, Proper, 1, End)) | EOS } } \\ \hline
{\tt 212 2nd Avenue } & {\tt 212-2nd-Avenue }  & \textcolor{darkgreen} {\tt 212-2nd-Avenue } \\
{\tt 124 3rd Avenue } & {\tt 124-3rd-Avenue }  & \textcolor{darkgreen} {\tt 124-3rd-Avenue } \\
{\tt 123 4th Avenue } & {\tt 123-4th-Avenue }  & \textcolor{darkgreen} {\tt 123-4th-Avenue } \\
{\tt 999 5th Avenue } & {\tt 999-5th-Avenue }  & \textcolor{darkgreen} {\tt 999-5th-Avenue } \\
\hline
{\tt 123 1st Avenue } & {\tt 123-1st-Avenue }  & \textcolor{darkgreen} {\tt 123-1st-Avenue } \\
{\tt 223 1stAvenue } & {\tt 223-1st-Avenue }  & \textcolor{red} {\tt 223-1stAvenue } \\
{\tt 112 2nd Avenue } & {\tt 112-2nd-Avenue }  & \textcolor{darkgreen} {\tt 112-2nd-Avenue } \\
{\tt 224 3rd Avenue } & {\tt 224-3rd-Avenue }  & \textcolor{darkgreen} {\tt 224-3rd-Avenue } \\
{\tt 123 5th Avenue } & {\tt 123-5th-Avenue }  & \textcolor{darkgreen} {\tt 123-5th-Avenue } \\
{\tt 99 5th Avenue } & {\tt 99-5th-Avenue }  & \textcolor{darkgreen} {\tt 99-5th-Avenue } \\
\hline
\end{tabular}
}}
\vspace{0.6cm}
\caption{Selected samples of incorrect model predictions on the Flashfill test set. These include both inconsistent programs, and consistent programs which failed to generalize.}
\label{fig:consistent_samples}
\end{figure}

\begin{figure}
\ContinuedFloat
\subfloat
{\centering 
\shiftleft{-10pt}{ 
\begin{tabular}{|p{0.34\linewidth}|p{0.34\linewidth}|p{0.22\linewidth}|}
\hline
\multicolumn{3}{ |p{0.9\textwidth}| } {\texttt{Model prediction: GetToken{\uscore}Word{\uscore}1 | Const(-) | GetToken{\uscore}Proper{\uscore}1(GetSpan(`;', -5, Start, `\#', 5, Start)) | GetUpto{\uscore}Comma Replace{\uscore}Space{\uscore}Dash | GetToken{\uscore}Word{\uscore}1(GetSpan(Proper, 4, End, `\$', 5, End)) | GetToken{\uscore}Number{\uscore}-5 | GetSpan(`\#', 5, End, `\$', 5, Start) | EOS }} \\
\hline
{\tt 28;\#DSI;\#139;\#ApplicationVirt{\hyphen}ualization;\#148;\#BPOS;\#138;\#Mi{\hyphen}crosoft PowerPoint } & {\tt DSI-ApplicationVirtualization-B{\hyphen}POS-Microsoft PowerPoint }  & \textcolor{red} {\tt DSI-Application } \\ \hdashline
{\tt 102;\#Excel;\#14;\#Meetings;\#55;{\hyphen}\#OneNote;\#155;\#Word } & {\tt Excel-Meetings-OneNote-Word }  & \textcolor{red} {\tt Excel-Meetings } \\ \hdashline
{\tt 19;\#SP Workflow Solutions;\#102;\#Excel;\#194;{\hyphen}\#Excel Services;\#46;\#BI } & {\tt SP Workflow Solut{\hyphen}ions-Excel-Excel Services-BI }  & \textcolor{red} {\tt SP Workflow Solutions-Excel } \\ \hdashline
{\tt 37;\#PowerPoint;\#141;\#Meetings;{\hyphen}\#55;\#OneNote;\#155;\#Word } & {\tt PowerPoint-Meetings-OneNote-Word }  & \textcolor{red} {\tt PowerPoint-Meetings } \\
\hline
{\tt 148;\#Access;\#102;\#Excel;\#194{\hyphen};\#Excel Services;\#46;\#BI } & {\tt Access-Excel-Excel Services-BI }  & \textcolor{red} {\tt Access-Excel } \\ \hdashline
{\tt 248;\#Bccess;\#102;\#Excel;\#194;{\hyphen}\#Excel Services;\#46;\#BI } & {\tt Bccess-Excel-Excel Services-BI }  & \textcolor{red} {\tt Bccess-Excel } \\ \hdashline
{\tt 28;\#DCI;\#139;\#ApplicationVirt{\hyphen}ualization;\#148;\#BPOS;\#138;\#{\hyphen}Microsoft PowerPoint } & {\tt DCI-ApplicationVirtualizat{\hyphen}ion-BPOS-Microsoft PowerPoint }  & \textcolor{red} {\tt DCI-Application } \\ \hdashline
{\tt 12;\#Word;\#141;\#Meetings;\#55;\#O{\hyphen}neNote;\#155;\#Word } & {\tt Word-Meetings-OneNote-Word }  & \textcolor{red} {\tt Word-Meetings } \\ \hdashline
{\tt 99;\#AP Workflow Solutions;{\hyphen}\#102;\#Excel;\#194;\#Excel Services;\#46;\#BI } & {\tt AP Workflow Solutions-Ex{\hyphen}cel-Excel Services-BI }  & \textcolor{red} {\tt AP Workflow Solutions-Excel } \\ \hdashline
{\tt 137;\#PowerPoint;\#141;\#Meetings;{\hyphen}\#55;\#OneNote;\#155;\#Excel } & {\tt PowerPoint-Meetings-OneNo{\hyphen}te-Excel }  & \textcolor{red} {\tt PowerPoint-Meetings } \\
\hline
\end{tabular}
}}

\vspace{0.6cm}
\caption{Selected samples of incorrect model predictions on the Flashfill test set. These include both inconsistent programs, and consistent programs which failed to generalize.}
\label{fig:consistent_samples}
\end{figure}

\end{document}